\documentclass[10pt,twocolumn,letterpaper]{article}

\usepackage{iccv}
\usepackage{times}
\usepackage{epsfig}
\usepackage{graphicx}
\usepackage{amsmath}
\usepackage{amssymb}
\usepackage{float}
\usepackage{stfloats}

\usepackage[pagebackref=true,breaklinks=true,letterpaper=true,colorlinks,bookmarks=false]{hyperref}

\iccvfinalcopy

\ificcvfinal\fi

\begin{document}

%%%%%%%%% TITLE
\title{Dehazing-NeRF: Neural Radiance Fields from Hazy Images} 

\author{Tian Li\\
Beihang University\\
% XueYuan Road No.37,HaiDian District,Beijing,China\\
{\tt\small 18374328@buaa.edu.cn}
\and
Lu Li\\
Beihang University\\
{\tt\small lilu@buaa.edu.cn}
\and
Wei Wang\\
Beijing Institute for General Artificial Intelligence\\
% First line of institution3 address\\
{\tt\small wangwei@bigai.ai}
\and
Zhangchi Feng\\
Beihang University\\
% First line of institution3 address\\
{\tt\small zcmuller@buaa.edu.cn}
}

\maketitle
\ificcvfinal\thispagestyle{empty}\fi

\begin{figure*}[t!]
\begin{center}
% \fbox{\rule{0pt}{2in} \rule{.9\linewidth}{0pt}}
\includegraphics[width=\textwidth]{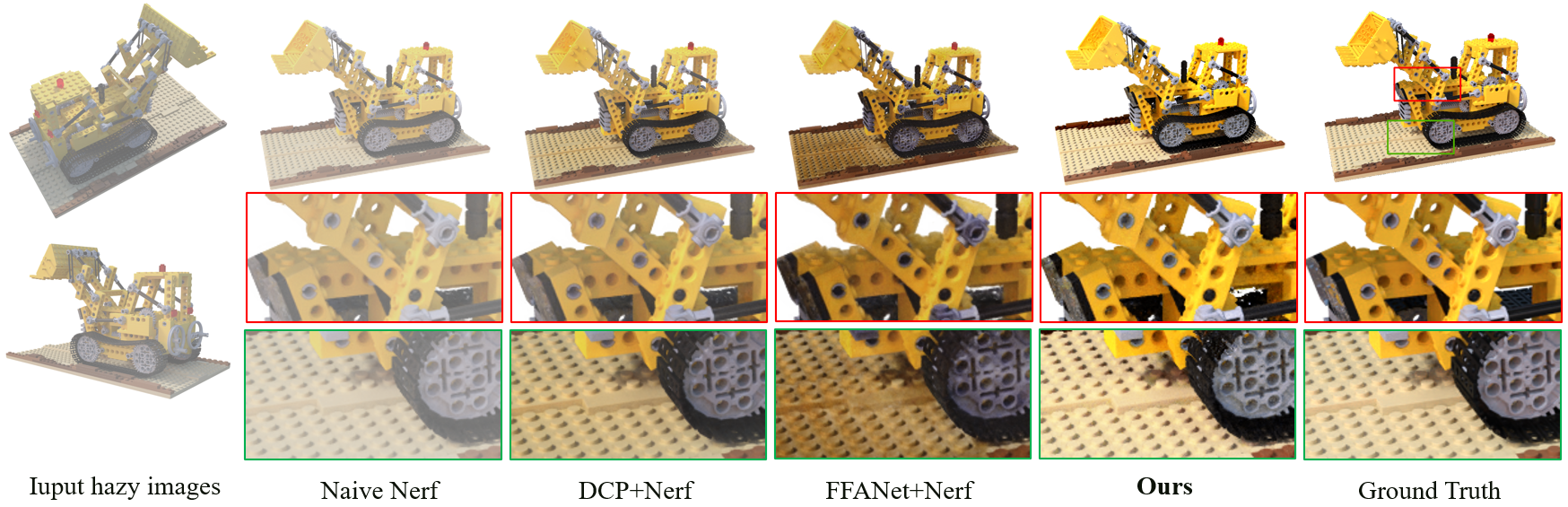}
\end{center}
   \caption{Comparison results on novel view image synthesis. Through comparing with several combinations of single-view dehazing methods and Nerf, e.g., DCP\cite{dcp}+Nerf and FFANet\cite{ffanet}+Nerf, Our Dehazing-NeRF can generate much more clear details by inferring atmospheric scattering parameters and learning Nerf model simultaneously. 
   %Given a set of hazy images, our dehazing-NeRF combined with the Neural Radiance Fields and restored parameters in the degeneration of hazy images. As shown in the figure, Dehazing-NeRF can synthesize clear novel view images with higher quality than the previous Baseline.
}
\label{fig:1}
\end{figure*}

%%%%%%%%% ABSTRACT
\begin{abstract}
   Neural Radiance Field (NeRF) has received much attention in recent years due to the impressively high quality in 3D scene reconstruction and novel view synthesis. However, image degradation caused by the scattering of atmospheric light and object light by particles in the atmosphere can significantly decrease the reconstruction quality when shooting scenes in hazy conditions. To address this issue, we propose Dehazing-NeRF, a method that can recover clear NeRF from hazy image inputs. Our method simulates the physical imaging process of hazy images using an atmospheric scattering model, and jointly learns the atmospheric scattering model and a clean NeRF model for both image dehazing and novel view synthesis. Different from previous approaches, Dehazing-NeRF is an unsupervised method with only hazy images as the input, and also does not rely on hand-designed dehazing priors. 
   %It only takes real hazy images as input and does not need real clear images. 
   % During model training, we employ a soft margin reconstruction consistency regularization, an atmospheric scattering consistency loss, a contrast discriminative loss and a total variation loss. 
   By jointly combining the depth estimated from the NeRF 3D scene with the atmospheric scattering model, our proposed model breaks through the ill-posed problem of single-image dehazing while maintaining geometric consistency. Besides, to alleviate the degradation of image quality caused by information loss, soft margin consistency regularization, as well as atmospheric consistency and contrast discriminative loss, are addressed during the model training process. 
   %and dehazing guidance loss to address quantization errors on hazy images, multi-view inconsistencies, and network convergence to trivial solutions. 
   Extensive experiments demonstrate that our method outperforms the simple combination of single-image dehazing and NeRF on both image dehazing and novel view image synthesis. 
   % Synthetic and real hazy scene data and code are available on GitHub.
\end{abstract}

%%%%%%%%% BODY TEXT
\section{Introduction}

Novel view synthesis and 3D geometric reconstruction are very important tasks in computer vision. The traditional explicit 3D representation methods mainly include 3D point cloud \cite{3dpoint}, voxel grid \cite{volume} and triangular patch \cite{mesh}. Recently, Neural Radiance Field (NeRF) \cite{mildenhall2021nerf} achieves photorealistic view synthesis with an implicit 3D representation. NeRF takes the spatial position and viewing direction of a 3D point as input, and predicts its radiance and volume density through a multi-layer perceptron, and then obtains the corresponding pixel value and depth through differentiable volume rendering \cite{diffvolume}.

Although NeRF has achieved good results in quite a few scenes, NeRF-based novel view image synthesis in hazy weather will deteriorate due to the invisibility of outdoor scenes~\cite{survey-dehaze}.
%areas , and the image data collected for novel view synthesis is affected by the haze weather.
%The imaging process of a hazy image is shown in Fig. \ref{}.
On foggy days, there are particles such as haze, fog, and water droplets in the atmosphere. These particles absorb the light reflected from the scene objects and scatter atmospheric light to form a degraded image. The degree of haze is positively correlated with the distance between the object and the observer. It is very difficult to model these particles in a NeRF model due to their unknown spatial distribution and physical property. 

%In addition, . As a result, the image degrades differently as the viewing direction changes, which creates inconsistencies across multiple viewpoints. This interferes with the novel view synthesis task.

%image enhancement-based dehazing algorithms are not designed for the dehazing task, often resulting in excessive dehazing.

%is not accurate enough for parameter estimation.

Traditionally, single-image dehazing is an ill-posed problem, which cannot estimate an accurate atmospheric scattering model~\cite{asm1976optics} (ASM) for an inverse solution. Some   priors on hazy images have been discovered to help restore image quality, e.g., dark channel prior~\cite{dcp}.
%often have the problem that the dehazing results do not match the real values. 
Deep supervised dehazing algorithms~\cite{ffanet,dehazeformer} usually need paired hazy and clean data for training, which is difficult to obtain in the real world.

In this paper, we propose a simple and effective Dehazing-NeRF model, which can not only recover a clear NeRF from hazy inputs for novel view synthesis, but also learn an atmospheric scattering model for image dehazing in an unsupervised manner. 
There are two branches in our model, one branch for estimating the parameters of ASM, namely atmosphere scattering coefficient $\beta$ and atmospheric light $A$, and the other branch for learning a hazy-free neural radiance field. Combining these two branches, ASM reconstructs a hazy image by fusing the rendered image, depth map, and atmospheric parameters.
During training, we propose a Soft Margin Reconstruction Consistency (SMRC) which handles the information losses during hazy image quantization, and several auxiliary losses to promote the convergence of model. Specifically, to ensure consistent atmosphere parameters across different viewpoints, we apply an Atmospheric Consistency Loss. Moreover, 
an image-based Contrast Discrimnative Loss between hazy image and the rendered image from the Novel view synthesis branch is introduced to learn a clear NeRF model. it is worth mentioning that our model does not rely on any hand-designed dehazing priors. As shown in Figure \ref{fig:1}, the experimental results demonstrate that our model can predict accurate atmosphere parameters in the unsupervised setting, and outperform the simple combination of single-image dehazing and NeRF on both image dehazing
and novel view image synthesis.

%During training, we jointly optimize the hazy image reconstruction branch and the NeRF branch, with only hazy images input as supervision, while during inference, the NeRF branch can synthesize clear rendering results from novel views.

%regularization to guide the reconstructed hazy image to be consistent with the real hazy image; we introduce a consistency loss to ensure that the parameter values estimated by different viewing angle images of the same scene are consistent; 

%We combine the atmospheric scattering model with NeRF, and by reconstructing the hazy image, DNN can accurately estimate the atmospheric scattering coefficient and airlight intensity. ASM obtains the reconstructed hazy image by fusing the clear image from the NeRF branch, the depth map, and the ASM parameters estimated by DNN. The NeRF branch takes uniformly sampled rays for the image position and orientation as input, and obtains an estimated clear image. 

Our contributions are summarized as follows:
\begin{itemize}
\item[$\bullet$] We propose an unsupervised novel view synthesis framework, which can recover clear NeRF from hazy inputs. We utilize the depth information of the 3D scene to supplement the undetermined parameter of ASM, which can solve the ill-posed problem of single-image dehazing.
% For hazy scenes, we proved its effectiveness. Our method does not require real clear images to provide supervision signals. 
\item[$\bullet$] We propose a soft margin consistency regularization to alleviate the problem of information loss caused by the quantization of hazy images while ensuring the consistency of reconstructed images. 
\item[$\bullet$] We experimentally validate that our method is capable of removing haze from images and synthesizing clear novel view images.
\end{itemize}

% \begin{figure}[t]
% \begin{center}
%    %\includegraphics[width=0.8\linewidth]{egfigure.eps}
% \end{center}
%    \caption{Example of a caption.  It is set in Roman so mathematics
%    (always set in Roman: $B \sin A = A \sin B$) may be included without an
%    ugly clash.}
% \label{fig:long}
% \label{fig:onecol}
% \end{figure}

% \begin{figure*}
% \begin{center}
% \fbox{\rule{0pt}{2in} \rule{.9\linewidth}{0pt}}
% \end{center}
%    \caption{Example of a short caption, which should be centered.}
% \label{fig:short}
% \end{figure*}

%------------------------------------------------------------------------
\section{Related work}
\label{related work}
Dehazing-NeRF combines concepts from several research fields. We review two main areas of related work: neural radiance fields and image dehazing.

\noindent\textbf{Neural Radiance Fields.} As a novel view synthesis and 3D reconstruction method, NeRF has been widely used in computer vision and graphics tasks, which uses coordinate-based implicit 3D scene representation. NeRF has recently received a lot of positive attention and has produced high-quality results in novel view synthesis assignments. Some extension work attempts to solve the problem of novel view synthesis under non-ideal image input, such as sparse view \cite{regnerf,dsnerf,infonerf,lolnerf}, blur \cite{deblurnerf,dpnerf}, low-resolution \cite{nerfsr,4KNeRF} image input. HDR-NeRF \cite{hdrnerf} and HDR-plenoxel \cite{hdrpleno}simulate the physical process of capturing images in the camera and high dynamic range radiance to solve the rendering problem of high-light ratio scenes.  By using truncated cone sampling, Mip-NeRF \cite{mipnerf} addresses the aliasing issue with light samples under multi-resolution image input. NeRF-W \cite{nerfw} introduces appearance and transient code to address the inconsistent appearance and transient objects in the gathered photos. By introducing fuzzy kernel estimation for NeRF, Deblur-NeRF solves motion blur and defocus blur. Although haze is very common, training NeRF with hazy images has not been explored, and the above algorithms do not consider the image degradation process caused by haze.

\begin{figure*}
\begin{center}
% \fbox{\rule{0pt}{2in} \rule{.9\linewidth}{0pt}}
\includegraphics[width=\textwidth]{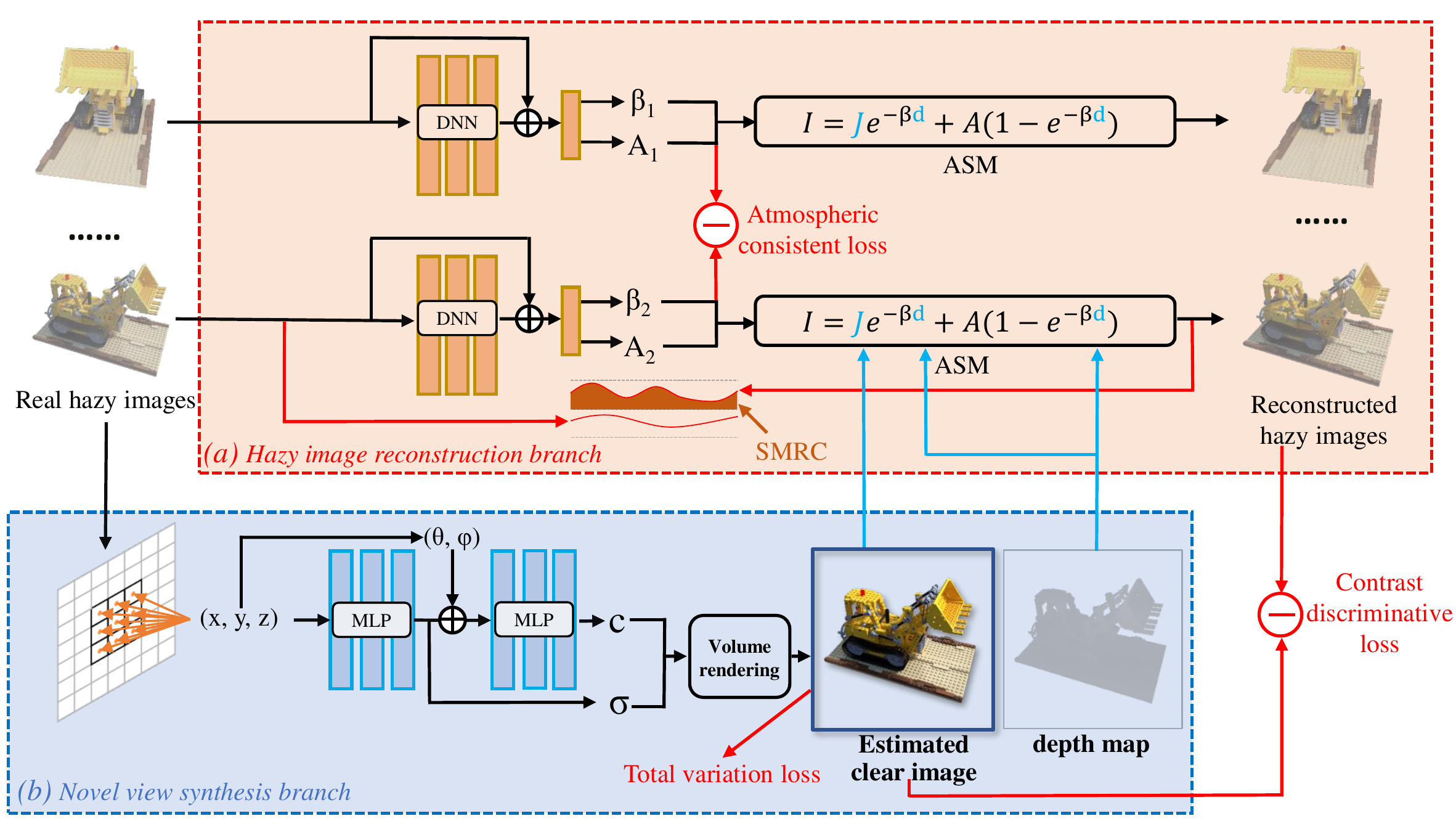}
\end{center}
   \caption{Overall pipeline for Dehazing-NeRF. (a) Hazy image reconstruction branch that estimates the parameter of the ASM. It combines clear image and depth maps from the Novel view synthesis branch to produce reconstructed hazy images. Where $A$ refers to the atmospheric light, $\beta$ is the scattering coefficient describing the scattering ability of the medium. (b) Novel view synthesis branch that generates estimated clear image $\hat{J}$.
}
\label{pipeline-fig}
\end{figure*}

\noindent\textbf{Image dehazing.} Image dehazing aims to recover high-quality clear scenes from hazy images. Image dehazing is a challenging task because it is an ill-posed problem. Before the widespread application of deep learning, image dehazing methods mainly relied on various prior assumptions and atmospheric scattering models \cite{asm}. For example, image dehazing based on dark channel priors \cite{dcp} is based on a priori assumption that in local patches of natural images that do not belong to the sky region, one of the pixel values of the three channels is always close to 0. The Color Attenuation Prior (CAP) \cite{cap} is based on the observation that differences in brightness intensity and image saturation make hazy regions of an image more variable than clear regions. The fog line theory \cite{hazeline} means that clusters of pixels from different depths in the scene will be connected into several straight lines (ie, fog lines) in RGB space. Also, some models \cite{dehazenet,aodnet,single,single2020} combine ASM with deep learning. Following this, deep learning-based end-to-end supervised dehazing algorithms \cite{selective,griddehazenet,ffanet,hierarchical,dehazeformer} demonstrated that non-atmospheric scattering models can also achieve comparable accuracy. These models are usually trained with a large number of pairs of hazy and clear images, aiming to Minimize the mean squared error bet we recently received a lot of positive attention o learn the relationship between the two. Semi-supervised dehazing methods \cite{an2022semi,chen2021psd,li2019semi,zhang2021single} can alleviate the requirement for pairs of hazy and clear images, which can utilize both labeled and unlabeled datasets. Since pairs of hazy and clear images are hard to obtain in the real world, some algorithms \cite{cong2020discrete,golts2019unsupervised,li2020zero} explore dehazing methods in an unsupervised manner. There are many difficulties in this task: (a) image dehazing is a ill-posed problem, and the unsupervised dehazing model is easy to converge to the trivial solution; (b) there is significant quantization loss in hazy images compared with clear images; (c) the ASM parameters estimated from a single image are not consistent from multiple views. This will interfere with novel view synthesis.

%------------------------------------------------------------------------
\section{Dehazing Neural Radiance Field}
In this section, we describe the process of reconstructing a clean NeRF given a set of hazy inputs. Figure \ref{pipeline-fig} shows the training pipeline of our approach. During this process, NeRF is optimized with the estimated clear image as a pseudo-supervision signal. Initially, we model the image degradation process due to haze, using an atmospheric scattering model as the physical scene prior (Section \ref{Pipeline}). To address the significant quantization loss present in hazy images, we then introduce soft margin reconstruction consistency (Section \ref{smrc}). Finally, we explain our loss function and optimization strategy to avoid network convergence to trivial solutions for training our model (Section \ref{Training}). It is worth noting that our model only takes real hazy images as input and does not require real clear images to provide supervision.
\subsection{Preliminary}

\noindent\textbf{Neural Radiance Field.}  NeRF~\cite{mildenhall2021nerf} implicitly defines a differentiable 3D scene, which takes 3D position $x$ and perspective direction $d$ as input, and takes volume density $\sigma$ And color $c$ as output. And it can approximate this mapping:
\begin{eqnarray}
(\mathbf{c}, \sigma)&=&F_{\boldsymbol{\Theta}}\left(\gamma_{L_{x}}(\mathbf{x}), \gamma_{L_{d}}(\mathbf{d})\right),
\end{eqnarray}
where $F_{\Theta}$ is the multi-layer perceptron with parameters $\Theta$, and $\gamma_{L}(\cdot)$ is the positional encoding that maps each element of a vector to a high-dimensional frequency space, $ \gamma_ {L_x} ( \cdot) $ and $ \gamma_ {L_x} (\cdot) $represents the encoding of position and direction respectively:
% \begin{small} 
\begin{eqnarray}
\!\gamma_{L}(x)\!\!=\!\left[\sin \pi x,\cos \pi x, \!\ldots\!, \sin 2^{L-1} \pi x,\! \cos 2^{L-1} \pi x\right]^{\mathrm{T}}\!,\!\!
\end{eqnarray}
% \end{small}
where $L$ is a hyperparameter that determines the frequency band. NeRF approximates volume rendering integrals using numerical quadrature to render the colors of rays passing through the scene. The color $C$ of the ray $\mathbf{r}$ can be approximated as:
% $R_{(t)} =o+td$ represents the ray passing through the optical center of the camera lens and the specified pixel of the image.
\begin{equation}
\hat{C}(\mathbf{r})=\sum_{i=1}^N T_i\left(1-\exp \left(-\sigma_i \delta_i\right)\right) \mathbf{c}_i,
\end{equation}
where $N$ is the number of 3D points sampled along the ray, $\delta_i$ and $\sigma_i$ are the predicted values of the color and volume density of the sampling point, $\delta_i$ is the distance between $(i+1)_{th}$ and $i_{th}$ sampling point, $T_i$ is the transmission factor, which represents the ray until the sampling point The probability of hitting any particle. $T_k$ is defined as:
\begin{equation}
T_k=\exp \left(-\sum_{l=1}^{k-1} \sigma_l \delta_l\right),
\end{equation}
where $\sigma_l$ is the volume density predicted by $F_\theta$ and $\sigma_l$ is the corresponding distance between adjacent orthogonal points. $k$ represents the number of layered sampling points between the near plane and the far plane of the camera. The parameters of the MLP are optimized by minimizing the sum of squared errors between the observed image pixel values and their estimated values:
\begin{equation}
\mathcal{L}=\sum_{i j}\left\|\mathbf{C}\left(\mathbf{r}_{i j}\right)-\hat{\mathbf{C}}\left(\mathbf{r}_{i j}\right)\right\|_2^2,
\end{equation}
where $\mathbf{C}\left(\mathbf{r}_{i j}\right)$ is the color of ray $j$ in image $I_i$.

\begin{figure}
\begin{center}
% \fbox{\rule{0pt}{2in} \rule{.9\linewidth}{0pt}}
\includegraphics[width=0.45\textwidth]{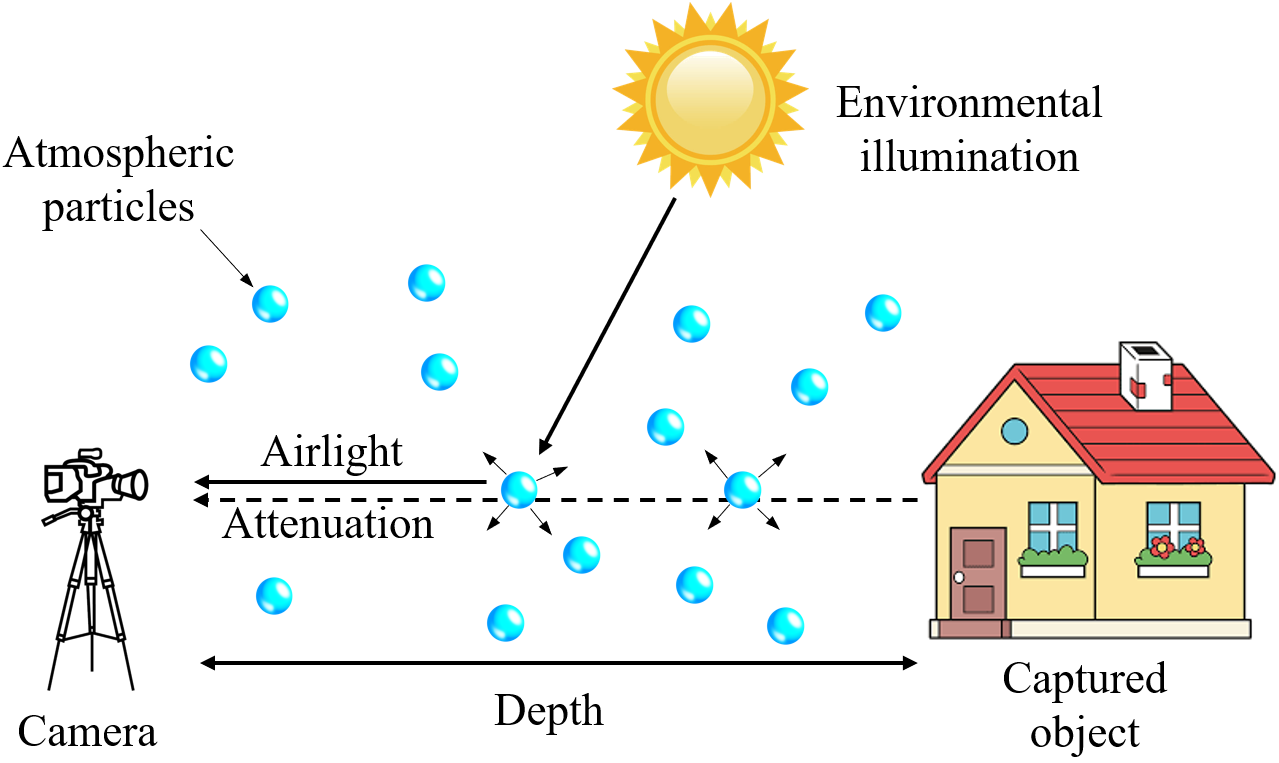}
\end{center}
   \caption{Schematic diagram of the atmospheric scattering model, same as \cite{guo2022image}.}
\label{fig:asm}
\end{figure}

\noindent\textbf{Atmospheric scattering model (ASM).}  As shown in Figure \ref{fig:asm}, the scattering effect of particles in the atmosphere is the main cause of haze, and the low visibility of hazy images is due to the absorption and scattering of light by suspended particles in the atmosphere \cite{asm,asm2000chromatic,asm2002vision}. The Atmospheric scattering model physically models the imaging process in hazy weather and analyzes the two reasons for image degradation in hazy weather. First, the light reflected by objects is absorbed and scattered by suspended particles and attenuated. Second, ambient light such as sunlight is suspended Particles scatter to form background light. ASM can be formulated as:
\begin{equation}
\begin{split}
I(x)&=J(x) t(x)+A(1-t(x)),\\
t(x)&=e^{-\beta d(x)}.
\end{split}
\end{equation}
where $I(x)$ is the observed hazy image, $J(x)$ denote the clear image to be recovered, $t(x)$ is the transmission map, $\beta$ is the scattering coefficient of the atmosphere, $A$ denotes the atmospheric light or airlight, and $d(x)$ is the distance between the object and the camera. Where $(x)$ represents pixel by pixel calculation. In the following description, we omit it for convenience.

\subsection{Pipeline}
\label{Pipeline}
As shown in Figure \ref{pipeline-fig}, Dehazing-NeRF contains two branches, the first branch is the hazy image reconstruction branch, and the second branch is the neural radiance field branch. During training, both branches are optimized in parallel. Its core idea is to reconstruct the estimated hazy image through the physical model, and in the process, get the parameters of ASM.

\noindent\textbf{Hazy image reconstruction branch.} 
The core of branch (a) is the physical model of image degradation. In this paper, ASM~\cite{asm} is used to reconstruct the degradation process of hazy images. There are three unknown parameters in ASM, which are atmospheric scattering coefficient $\beta$, atmospheric light intensity $A$ and depth map $\hat{D}$. To solve the undetermined problem of image dehazing and reduce the number of parameters to be estimated, we use the depth image $\hat{J}$ predicted by branch (b), which simultaneously guarantees the geometric consistency of the reconstructed clear scene. In order to deal with the problem of few samples, branch (a) uses the pre-trained DNN to predict the two parameters of $\beta$ and $A$. From the above parameters and the clear image estimated by branch (b), the estimated hazy image $\hat{I}$ is reconstructed using ASM. Theoretically, the reconstructed hazy image should be consistent with the real hazy image $I$. 
% In general, the input of branch (a) is the real hazy image, the clear image $\hat{J}$ and the depth map $\hat{D}$ estimated by branch (b), and the output is the reconstructed hazy image $\hat{I}$.

\noindent\textbf{Novel View Synthesis Branch. }
The input of branch (b) is the direction and position of the corresponding viewing angle, and the output is the estimated clear image $\hat{J}$. The direction and position of the viewing angle correspond to the real hazy image $I$ input by branch (a). Different from the random sampling of the original NeRF~\cite{mildenhall2021nerf} in the whole image, the real hazy image input by branch (a) of this framework has been down-sampled at equal intervals, and the sampling offset is randomly generated. This can ensure that the sampling results are aligned with the image, and then use the pre-trained model on a large-scale data set to avoid the difficulty in estimating the parameters in ASM due to the small number of samples. At the same time, down-sampling reduces the GPU memory consumed by the entire framework.

The optimization process of this framework is detailed in Section \ref{Training}. In general, we hope that the estimated hazy image is consistent with the real hazy image, and the parameter values estimated from different viewing angle images are as consistent as possible, while the estimated clear image should not contain haze.

\subsection{Soft Margin Reconstruction Consistency}
\label{smrc}

\begin{figure}
\begin{center}
% \fbox{\rule{0pt}{2in} \rule{.9\linewidth}{0pt}}
\includegraphics[width=0.45\textwidth]{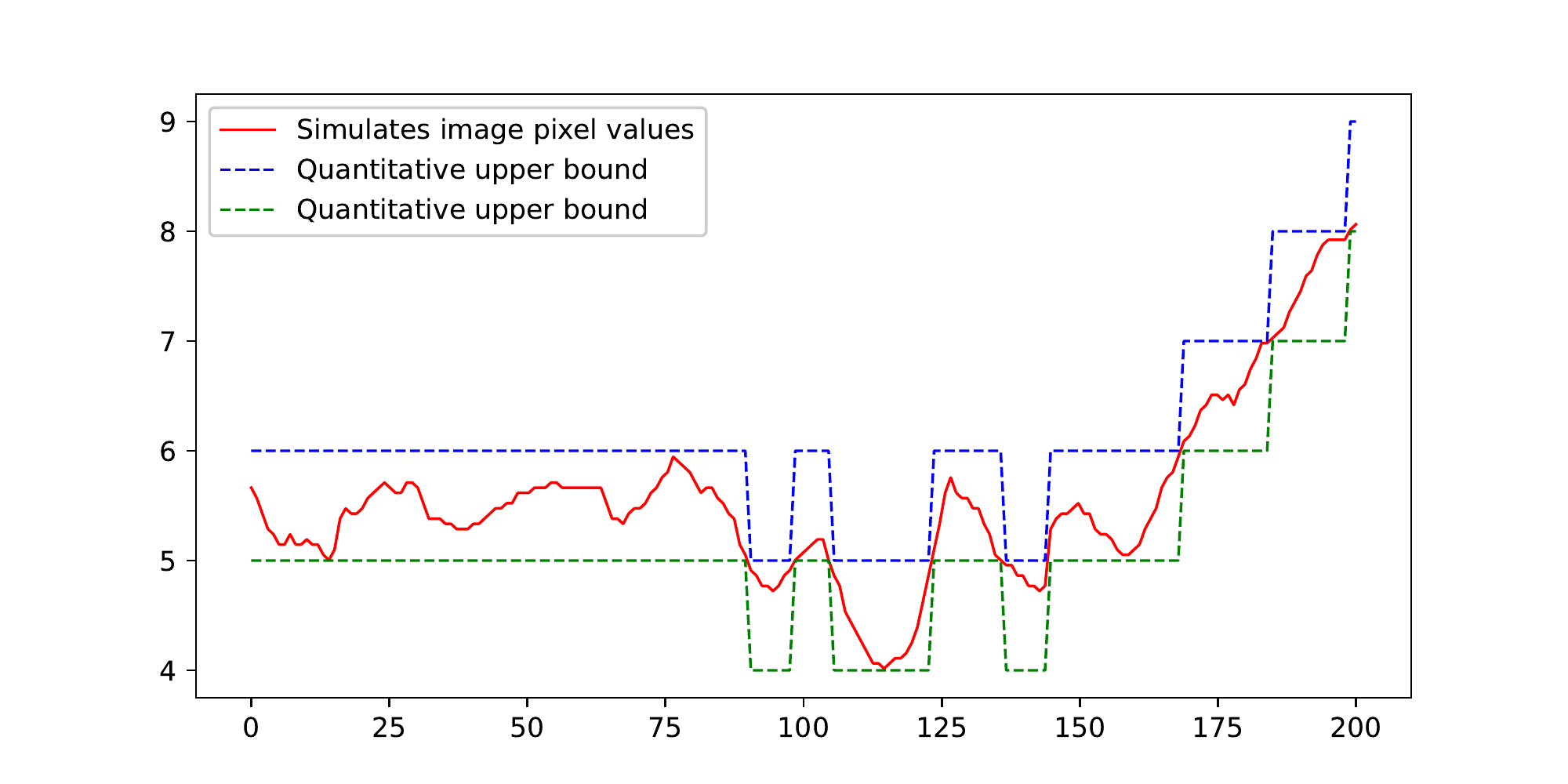}
\end{center}
   \caption{Quantization loss of images. After the simulated image is quantized, some information is lost. This is especially acute in hazy images where the image histogram is compressed.}
\label{fig:quan}
\end{figure}

There is a huge difference in the pixel histograms of the hazy image and the hazy-free image, and the pixel values of the hazy image are often compressed to a smaller range~\cite{survey-comprehensive}. The quantization series of digital images is usually 256, and the quantization series corresponding to most pixel values of hazy images are greatly reduced, so that a considerable amount of information in the image has been lost (Figure \ref{fig:quan}). while The image pixel values in different viewing directions are different, which ultimately makes it difficult for NeRF to converge, and the rendering results of novel view synthesis often have obvious quantization noise, shown as Figure \ref{fig:closs}.

To cope with this problem, we propose a soft-margin consistency regularization. Quantization assigns the signal $q$ not to a single value, but to an interval $(bi, b_{i+1})$. Inspired by previous work~\cite{quanti}, we do not strictly constrain the real image to be exactly the same as the reconstructed image, but allow a slight difference between the two, which is determined by the size of the quantization level. The reconstruction regularization can be formulated as:

\begin{equation}
\mathcal{L}_{\text {rec}}=\left\|I-\hat{I}\right\|_{SMRC},
\end{equation}
where I is the real hazy image, and $\hat{I}$ is the reconstructed hazy image, $\left\|\cdot \right\|_{SMRC}$is the proposed soft margin reconstruction consistency regularization. It can be formulated as:
\begin{equation}
\left\|f(u, b)\right\|_{SMRC}=\left\{\begin{array}{ll}
\lambda \left(u-b\right)^{2}, & \text { if } b_{i}<=u<=b_{i+1} \\
\left(u-b_{i}\right)^{2}, & \text { if } u<b_{i} \\
\left(u-b_{i+1}\right)^{2}, & \text { if } u>b_{i+1}
\end{array}\right.
\end{equation}
where $u$ is the predicted value of the pixel, and $b$ is the real pixel value, $(b_i, b_{i+1})$is the quantization interval, and $\lambda$ is a coefficient between 0 and 1.

\subsection{Auxiliary loss function}
\label{Training}
% There are many difficulties in this task: (a) image dehazing is an undetermined problem, and the model is easy to converge to the trivial solution; (b) there is a significant quantitative loss in the hazy image compared with the clear image; (c) ASM parameters estimated from a single image are not consistent from multiple perspectives; 
In order to solve the problems mentioned in section \ref{related work}, we carefully designed the following auxiliary loss functions to alleviate the problems of noise in the reconstruction results, model convergence to trivial solutions, and inconsistency in multiple views.

\noindent\textbf{Atmospheric consistency loss.} It is used to ensure that the atmosphere parameter values estimated by different view images of the same scene are consistent, which relies on the prior with the static hazy scene assumption. It helps the convergence of the NeRF and avoids the loss of detail in hazy scenes. It is defined as:
\begin{equation}
\mathcal{L}_{cons} =\frac{\sum_{i=1}^{N}\left(\beta _{i}-\bar{\beta} \right)^{2}+\left(A_{i}-\bar{A} \right)^{2}}{N},
\end{equation}
where $\beta_{i}$ and $A_{i}$ are the atmospheric scattering coefficient and atmospheric light intensity of the $i_{th}$ image estimated by DNN, respectively. N is the number of multi-view images. 

\noindent\textbf{Contrast discriminative loss. }It is used to ensure the difference between hazy and clear images, avoiding trivial solutions. It is defined as:
\begin{equation}
\mathcal{L}_{cd} = Lc(I)-Lc(\hat{J}),
\end{equation}
where $Lc(\cdot)$ is used to calculate the local contrast of the image. It can be formulated as:

\begin{equation}
Lc=\left\|I-Upsample(Avgpool(I))\right\|_{2},
\end{equation}

\noindent where $I$ represents the input image, $upsample(\cdot)$ represents upsampling, and $Avgpool(\cdot)$ represents the mean pooling, where the upsampling magnification is the same as the step size and kernel size of the mean convolution. 

\noindent\textbf{Total Variation loss.} In order to avoid the noise of novel view synthesis result and ensure its smoothness, we add the total variation loss~\cite{tvloss} to constrain the gradient of image pixel value:
\begin{equation}
\mathcal{L}_{tv}=\int_{\Omega_{u}}\left|\nabla_{u}\right| d x d y=\int_{D_{u}} \sqrt{u_{x}^{2}+u_{y}^{2}} d x d y,
\end{equation}

\noindent where $D_u$ is the neighborhood for images, and $u_{x}=\frac{\partial u}{\partial x}, u_{y}=\frac{\partial u}{\partial y}$. The overall loss function of this framework is the sum of SMRC, Atmospheric consistency loss, Contrast discrimination loss and Total Variation loss, namely:
\begin{equation}
\mathcal{L}_{tot} =\mathcal{L}_{rec}+\lambda_{1}\mathcal{L}_{cons}+\lambda_{2}\mathcal{L}_{cd}+\lambda_{3}\mathcal{L}_{tv},
\end{equation}
where $\lambda_{1}$, $\lambda_{2}$ and $\lambda_{3}$ are coefficients used to balance different losses. $\mathcal{L}_{rec}$, $\mathcal{L}_{cons}$ and $\mathcal{L}_{cd}$ are soft margin reconstruction consistency, Atmospheric consistency loss, contrast discriminative loss and total variation loss, respectively.

\section{Experiment}
\begin{table*}[]
\footnotesize
\centering
\begin{tabular}{c|ccc|ccc|ccc|ccc}
\hline
                   & \multicolumn{3}{c}{Lego}                          & \multicolumn{3}{c}{Chair}                         & \multicolumn{3}{c}{Drums}                         & \multicolumn{3}{c}{Ficus}                          \\
Method             & PSNR           & SSIM            & LPIPS           & PSNR           & SSIM            & LPIPS           & PSNR           & SSIM            & LPIPS           & PSNR           & SSIM            & LPIPS           \\ \hline
Naive NeRF~\cite{mildenhall2021nerf}         & 16.67          & 0.8800          & 0.0740          & 22.00          & 0.9418          & 0.0459          & 18.12          & 0.8896          & 0.0910          & 21.94          & 0.9480          & 0.0429          \\
DCP~\cite{dcp}+NeRF           & 23.71          & \textbf{0.9299} & \textbf{0.0333} & 26.73          & \textbf{0.9659} & \textbf{0.0251} & 21.68          & 0.9009          & 0.0743          & 24.43          & 0.9537          & 0.0382          \\
FFANet~\cite{ffanet}+NeRF        & 20.73          & 0.8979          & 0.0543          & 23.88          & 0.9386          & 0.0500          & 22.00          & 0.8893          & 0.0835          & \textbf{27.49} & \textbf{0.9592} & \textbf{0.0332} \\
Dehaze-former~\cite{dehazeformer}+NeRF & 22.25          & 0.9134          & 0.0451          & 22.47          & 0.9441          & 0.0411          & 21.53          & \textbf{0.9013} & \textbf{0.0671} & 25.78          & 0.9584          & 0.0345          \\ \hline
Dehazing-NeRF (Ours)               & \textbf{26.61} & 0.8997          & 0.0606          &\textbf{28.80}  &  0.9451        &0.383         & \textbf{23.23} & 0.8876      & 0.0873        &  27.26        & 0.9541         & 0.0399          \\ \hline
\end{tabular}
\caption{
Quantitative results in synthetic hazy scenes. The experimental results demonstrate that our method delivers competitive performance compared to prior works.}
\label{compar-syn}
\end{table*}

\begin{table}[]
\footnotesize
\centering
\begin{tabular}{c|ccc}
\hline
Scene & $\beta$ & $A$    & Average relative error \\ \hline
Lego  & 0.1388  & 0.9309 & 15.34\%                \\
Chair & 0.1548  & 0.9080 & 8.97\%                 \\
Drums & 0.1983  & 0.8522 & 14.47\%                \\
Ficus & 0.1397  & 1.0793 & 24.34\%                \\ \hline
GT    & 0.162   & 0.8    &                        \\ \hline
\end{tabular}
\caption{Parameter value prediction results $\hat{\beta}$ and $\hat{A}$ for different scenes. Without relying on the real parameter value for training, the parameter value estimated by our model is close to the ground truth.}
\label{tab:lhbeta}
\end{table}

\subsection{Implementation Details.} 

We base our unsupervised dehazing model on a Pytorch re-implementation of NeRF~\cite{nerfacc}.We also use 20k iterations to train each scene. We use the Adam optimizer~\cite{adam} with default parameters. In the neural radiance field model and the DNN model, we adopt the MultiStepLR optimizer, and the initial learning rate is adjusted to 1e-2 and 3e-4, respectively. At 1/3 times, 3/5 times, 4/5 times, and 9/10 times the maximum number of iterations, the learning rate is reduced by a factor of 0.33. We train each scene for 20k iterations on an NVIDIA TITAN RTX GPU. We adopt the same MLP structure as the original NeRF~\cite{mildenhall2021nerf}, and for DNN, we initially adopt the pre-trained ResNet18~\cite{he2016deep} model to improve the adaptability to few samples.

During training, we jointly optimize branch (a) and branch (b) to ensure static scene priors and geometric consistency. While in the inference stage, we only compute branch (b), and by inputting different viewing angles and observation positions, we get clear images from novel viewing angles.

\subsection{Datasets and Metrics.} 
\noindent\textbf{Benchmark datasets.} 
To evaluate the performance of our network, we have made a new novel view synthetic data set for hazy scenes based on the previous synthetic data sets, which are based on the Realistic Synthetic 360$^{\circ}$ dataset~\cite{mildenhall2021nerf}, which is generated from 8 virtual scenes. We assume that there is uniform haze in the scene.  We haze the original clear image according to ASM. For synthetic scenes, we adopt the true depth of the test set. In order to facilitate comparison, we have added the same concentration of haze to different scenarios. It is worth noting that for Realistic Synthetic 360$^{\circ}$, we add the image of the original test set as a training set, and the original training set and verification set are used as a test set. The posture and internal parameters of the camera are consistent with the original dataset.

% \subsection{Baseline methods and evaluation metrics.} 
\noindent\textbf{Baseline methods and evaluation metrics.}
Due to the lack of research on the combination of novel view synthesis and dehazing, we carefully selected several methods for comparison. The first is to directly use hazy inputs to train NeRF, and the remaining methods are representative traditional methods and several state-of-the-art methods learned-based dehazing methods, namely FFA-Net~\cite{ffanet}, Dehaze-former~\cite{dehazeformer} and the dark channel prior~\cite{dcp} dehazing method of He et al. Because single-image dehazing algorithms are not suitable for novel view image synthesis, we first apply them to each hazy image to obtain recovered images, and then feed these dehazing results into NeRF for training.
The quality of the rendered image is evaluated using metrics commonly used in image augmentation and NeRF, namely the PSNR, SSIM and LPIPS~\cite{lpips} metrics between the rendered image and the ground truth clear image.

\begin{figure*}
\begin{center}
% \fbox{\rule{0pt}{2in} \rule{.9\linewidth}{0pt}To
\includegraphics[width=\textwidth]{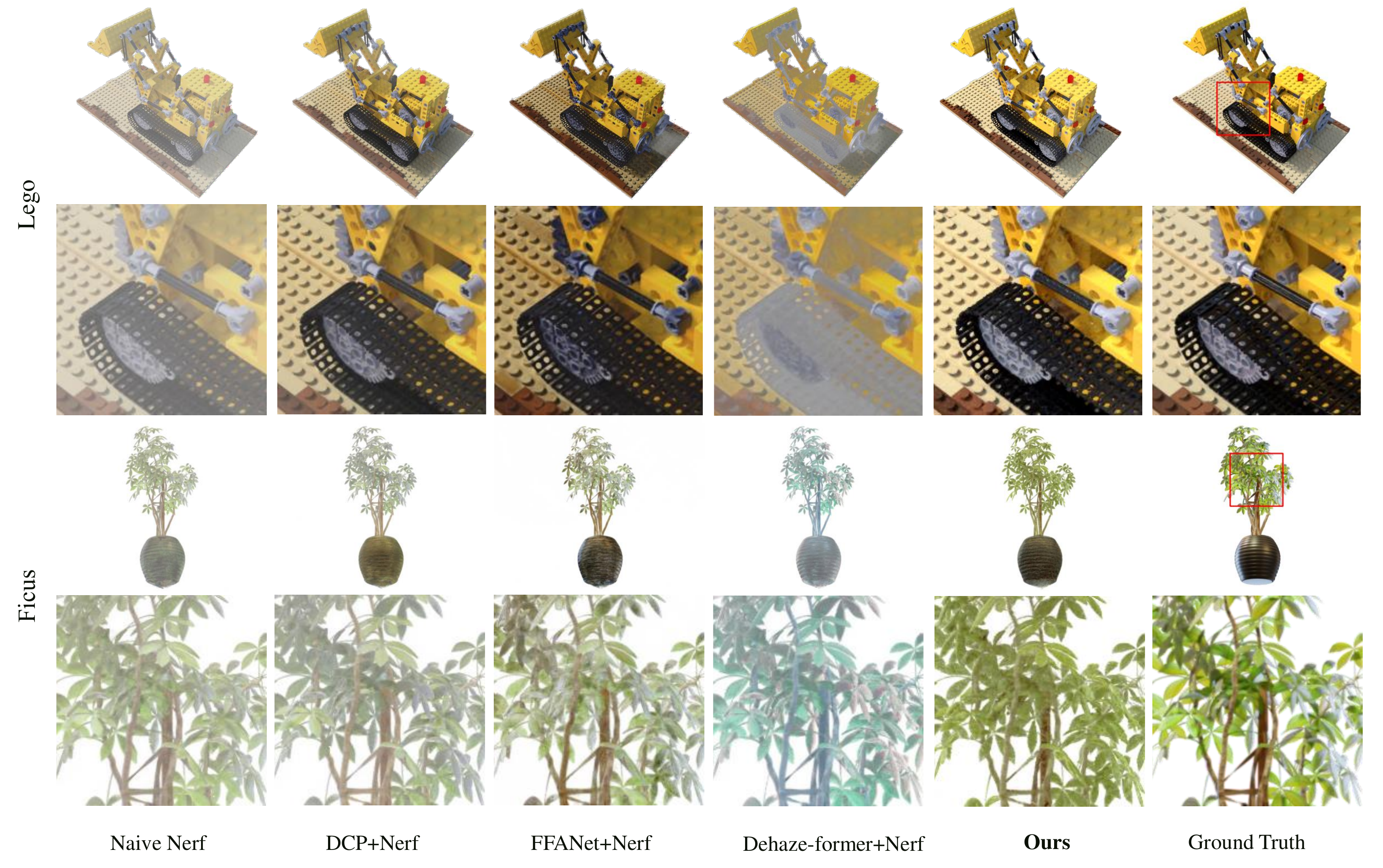}
\end{center}
   \caption{Qualitative comparison on hazy images input. As shown in the figure, Dehazing-NeRF can synthesize clear novel view images with higher quality than the previous Baseline.
}
\label{fig:comp4}
\end{figure*}
% \subsection{Novel View Synthesis}

\subsection{Comparisons}
\noindent\textbf{Comparison with baseline methods. }Since there is no work on reconstructing NeRF from hazy inputs, we prudently select several possible baselines for comparison. The easiest way is to train NeRF directly with hazy inputs (naive NeRF~\cite{mildenhall2021nerf}). In addition, we compare with baselines in image space, First, we use existing image dehazing algorithms to remove haze, and then training NeRF on dehazed images. We compare current classical (DCP~\cite{dcp}+NeRF) and state-of-the-art methods for single-image dehazing (Dehaze-former~\cite{dehazeformer}+NeRF). We present quantitative results on synthetic scenes in Table \ref{compar-syn}. Our entire pipeline outperforms these baselines in synthetic scenes, Figure \ref{compar-syn} and Figure \ref{fig:comp4} show qualitative comparison results with hazy image inputs, respectively. Our method's novel view synthesis results are the closest to the ground truth in overall color and free from artifacts around objects. 

% The PSNR of our model is relatively high, but the local structure metrics such as SSIM and LPIPS~\cite{lpips} are relatively low. The reason may be that the supervised deep learning model contains implicit filtering process, while DCP has designed the displayed filtering process, which alleviates the problem of high frequency noise. It is worth noting that FFA-Net~\cite{ffanet} and Dehaze-former~\cite{dehazeformer} have conducted long-term training on the RESIDE OTS data set~\cite{ots}, which contains 313950 paired images, while our model input only contains 200 hazy images for each scene.

The overall photometric index PSNR of our model is relatively good, but the local structure metrics (such as SSIM and LPIPS~\cite{lpips}) are relatively weak. Our model performs better on the overall metrics, probably because our model estimates the haze generation process better. The reason for the weaker local structural metrics may be that these local metrics are more sensitive to noise. While supervised deep learning models contain an implicit filtering process, DCP explicitly designs the filtering process, which alleviates the problem of high-frequency noise. It is worth noting that FFA-Net~\cite{ffanet} and Dehaze former~\cite{dehazeformer} have been trained on the RESIDEOTS dataset ~\cite{OTS}, which contains 313950 paired images, while our model input contains only 200 blurred images per scene. Our model can still achieve comparable results under the premise that the training data is several orders of magnitude less than the comparison methods.

\begin{figure}
\begin{center}
% \fbox{\rule{0pt}{2in} \rule{.9\linewidth}{0pt}To
\includegraphics[width=0.45\textwidth]{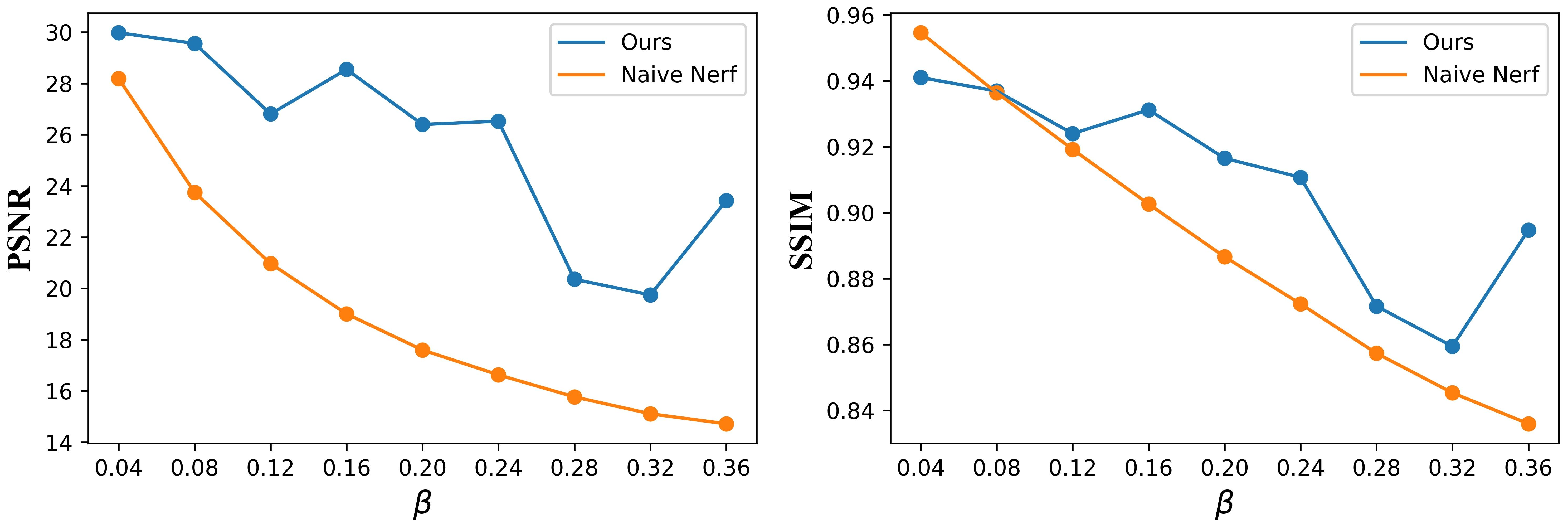}
\end{center}
   \caption{Quantitative results at different haze concentrations in Lego scene. As the atmospheric scattering coefficient $\beta$ increases, the results of Naive-NeRF~\cite{mildenhall2021nerf} are greatly affected, while our results are more robust to different haze concentrations.
}
\label{betas}
\end{figure}

\noindent\textbf{Comparison of different haze concentrations.}
We added different concentrations of haze to the Lego scene according to ASM. Specifically, we set $A$ to 0.8, and the atmospheric scattering coefficient $\beta$ increased from 0.04 to 0.36. As shown in Figure \ref{betas}, in Lego scene with different levels of haze, the results of Naive-NeRF~\cite{mildenhall2021nerf} are greatly affected, which indicates that the haze affects the 3D reconstruction to a large extent, and our model has a large impact on the different degrees of Haze has strong robustness.

\noindent\textbf{Estimation of parameters in ASM.}
As shown in Table \ref{tab:lhbeta}, in the four synthetic scenes, our model estimates the parameter values in ASM more accurately, which also shows the effectiveness of Dehazing-NeRF in modeling the image degradation process due to haze. The reason for the biased estimation of the parameter values in this model may be that the data lacks ground truth of parameters.
% \subsection{Novel View Synthesis}

\subsection{Ablation Study}
We conduct ablations on several constraints in our framework: soft margin reconstruction consistency (w/o SMRC), atmospheric consistency loss, contrast discrimination loss, and total variation loss. We remove these constraints individually and train NeRF in a Lego scene. We report quantitative metrics of synthesized novel views and ground truth novel views. In the absence of contrast discrimination loss, the novel view synthesis results fail to remove the haze, and without SMRC, the framework cannot reconstruct the 3D scene at all. As shown in Table \ref{tab:ablation}, overall, the best results are obtained when using full constraints.

\begin{figure}
\begin{center}
\includegraphics[width=0.5\textwidth]{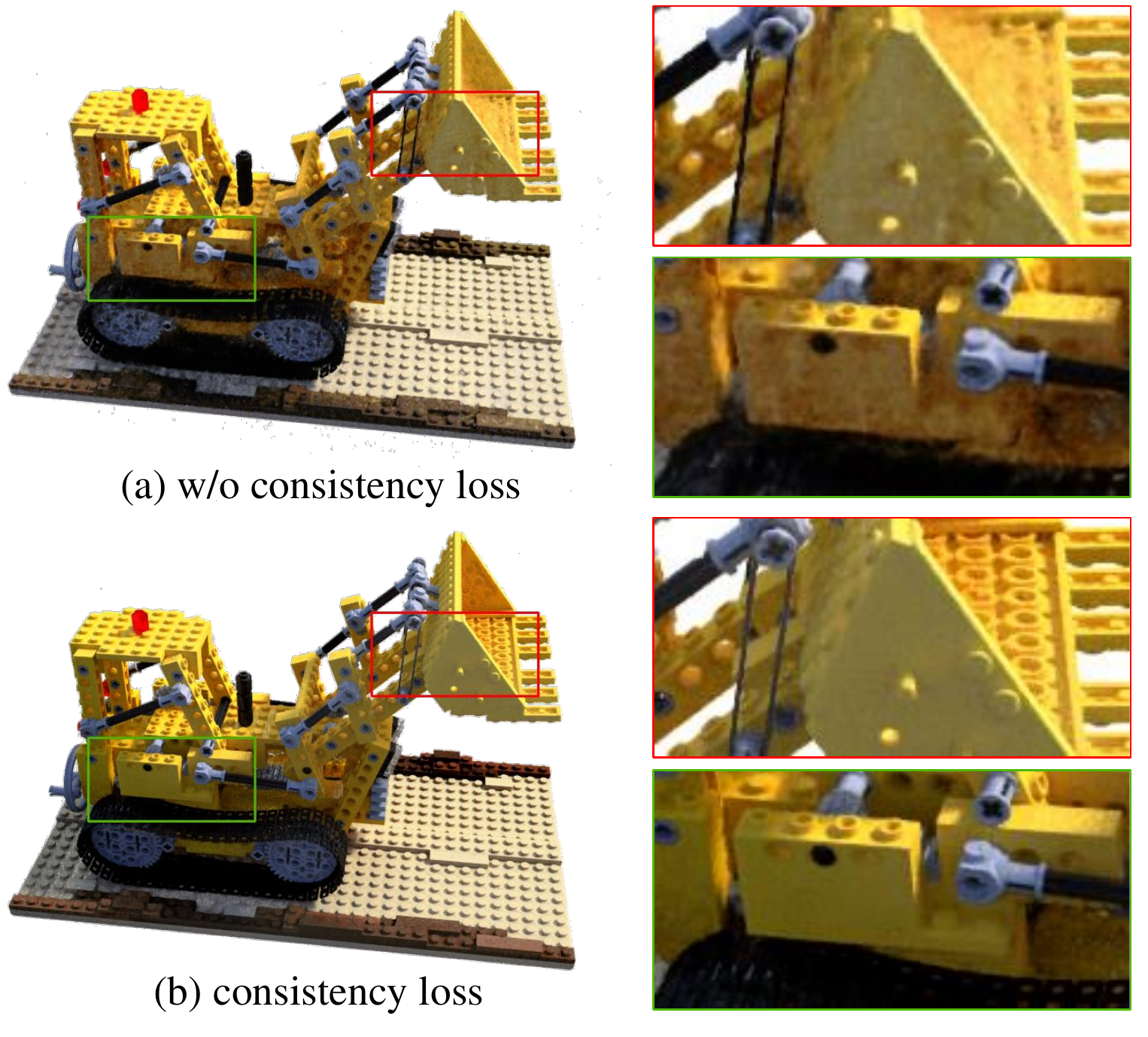}
\end{center}
   \caption{Examples of ablations in Lego scene. Figure (b) has more realistic details and less quantization noise than Figure (a). Note the Local details highlighted in the boxes.
}
\label{fig:closs}
\end{figure}

We separately remove the atmospheric consistency loss and compare its impact on the reconstruction quality. We show the qualitative results of Lego scene in Figure \ref{fig:closs}. We note that after adding the atmospheric consistency loss, the local detail reconstruction results of the scene are better. Local bumps on the bucket of the bulldozer have a better performance. Near the track, there is less noise and the local structure is closer to the ground truth. This may be because when the parameters $\beta$ and $A$ estimated by DNN for different view images are inconsistent, the multi-view consistency from different views becomes worse, which in turn interferes with the optimization of 3D scenes.

\begin{table}[]
\footnotesize
\centering
\begin{tabular}{l|ccc}
\hline
Method & PSNR & SSIM  & LPIPS  \\ \hline
Ours  & \textbf{26.61}   & \textbf{0.8997}  &  \textbf{0.0606}             \\
-w/o SMRC &  3.55 &  0.3057  &  0.7350          \\
-w/o Atmospheric consistency loss &  22.25 &  0.8362  &  0.0827  \\
-w/o Contrast discrimination loss & 16.36  &  0.8578 &    0.0949        \\ 
-w/o Total Variation loss    & 22.71  & 0.8404    &    0.0842               \\ \hline
\end{tabular}
\caption{Quantitative results of ablation analysis on photometric errors of Lego scene, which shows that the combination of all the designed components is the best.}
\label{tab:ablation}
\end{table}

\section{Discussion and Conclusion}
\subsection{Discussion}
Blindly restoring clear 3D scenes and parameters in ASM~\cite{asm} only from hazy images at the same time is an ill-posed problem. NeRF~\cite{mildenhall2021nerf} can also reconstruct a hazy scene, which can be matched to the input hazy image. So how does our framework able to get a clear NeRF? We also encountered this problem during the experiment, the model predicted that the atmospheric scattering coefficient in ASM was 0, which caused the NeRF reconstruction to remain hazy. However, the degree of haze in the image varies with depth, and there is a positive correlation between the two, and the hazy 3D scene is not enough to be consistent with the real hazy image from different views. Even so, NeRF tends to converge to trivial solutions. We introduce the contrast discriminative loss, which can avoid blurring results by ensuring that there is a difference between the estimated hazy image and the estimated hazy-free image. Our framework makes use of the depth map rendered by NeRF, which complements the ASM undetermined parameter. By imposing enough constraints on our framework, it becomes possible to restore a clear nerf only from hazy images.

In addition, an another feasible method for this task is to directly use NeRF to model the hazy scene, and then explicitly remove the haze by “subtracting” small density values in the voxel mesh. However, this idea has been found to be problematic in practice, that is, white haze is often modeled intensively to the surface of the object, and the color of the surface is mixed with the white of the haze, and it is impossible to simply distinguish between the two.
\subsection{Limitations and future work}
One of the main shortcomings of Dehazing-NeRF is that it relies on COLMAP~\cite{colmap,colmap2} to restore the camera's pose information, which may fail when the haze is very severe, but some recent work has tried to jointly optimize NeRF and camera poses. In addition, using Dehazing-NeRF to restore NeRF from images with quantization loss tends to give results with high-frequency noise. This is due to the loss of information during image quantization. Restoring a clear 3D scene from a sequence of hazy images is an ill-posed problem. Because of the lack of real hazy-free images, it is often difficult for unsupervised methods to obtain better results than supervised learning. 
\subsection{Conclusion}
This paper proposes Dehazing-NeRF, a unified novel view synthesis pipeline to achieve photorealistic rendering from hazy inputs and without haze-free data. Specifically, we take advantage of the 3D consistency in NeRF and combine NeRF and ASM, supplement the undetermined parameters in ASM, and use the atmospheric consistency loss and Contrast discrimination loss to guide the model to realize the reconstruction of hazy images, thus Clear novel view synthesis results are achieved. Experiments hazy scenes validate the effectiveness of our framework, showing significant improvements in quality compared to image-space methods. We hope that our work can advance the development of NeRF and image dehazing, making NeRF robust to non-ideal inputs.

{\small
\bibliographystyle{ieee_fullname}
\bibliography{egbib}

\begin{thebibliography}{10}\itemsep=-1pt

\bibitem{an2022semi}
Shunmin An, Xixia Huang, Le Wang, Linling Wang, and Zhangjing Zheng.
\newblock Semi-supervised image dehazing network.
\newblock {\em The Visual Computer}, 38(6):2041--2055, 2022.

\bibitem{mipnerf}
Jonathan~T. Barron, Ben Mildenhall, Matthew Tancik, Peter Hedman, Ricardo
  Martin-Brualla, and Pratul~P. Srinivasan.
\newblock Mip-nerf: A multiscale representation for anti-aliasing neural
  radiance fields, 2021.

\bibitem{hazeline}
Dana Berman, Shai Avidan, et~al.
\newblock Non-local image dehazing.
\newblock In {\em Proceedings of the IEEE conference on computer vision and
  pattern recognition}, pages 1674--1682, 2016.

\bibitem{dehazenet}
Bolun Cai, Xiangmin Xu, Kui Jia, Chunmei Qing, and Dacheng Tao.
\newblock Dehazenet: An end-to-end system for single image haze removal.
\newblock {\em IEEE Transactions on Image Processing}, 25(11):5187--5198, 2016.

\bibitem{chen2021psd}
Zeyuan Chen, Yangchao Wang, Yang Yang, and Dong Liu.
\newblock Psd: Principled synthetic-to-real dehazing guided by physical priors.
\newblock In {\em Proceedings of the IEEE/CVF conference on computer vision and
  pattern recognition}, pages 7180--7189, 2021.

\bibitem{cong2020discrete}
Xiaofeng Cong, Jie Gui, Kai-Chao Miao, Jun Zhang, Bing Wang, and Peng Chen.
\newblock Discrete haze level dehazing network.
\newblock In {\em Proceedings of the 28th ACM International Conference on
  Multimedia}, pages 1828--1836, 2020.

\bibitem{mesh}
Nico Cornelis, Bastian Leibe, Kurt Cornelis, and Luc Van~Gool.
\newblock 3d urban scene modeling integrating recognition and reconstruction.
\newblock {\em International Journal of Computer Vision}, 78:121--141, 2008.

\bibitem{3dpoint}
Yasutaka Furukawa and Jean Ponce.
\newblock Accurate, dense, and robust multiview stereopsis.
\newblock {\em IEEE transactions on pattern analysis and machine intelligence},
  32(8):1362--1376, 2009.

\bibitem{golts2019unsupervised}
Alona Golts, Daniel Freedman, and Michael Elad.
\newblock Unsupervised single image dehazing using dark channel prior loss.
\newblock {\em IEEE transactions on Image Processing}, 29:2692--2701, 2019.

\bibitem{survey-comprehensive}
Jie Gui, Xiaofeng Cong, Yuan Cao, Wenqi Ren, Jun Zhang, Jing Zhang, Jiuxin Cao,
  and Dacheng Tao.
\newblock A comprehensive survey and taxonomy on single image dehazing based on
  deep learning.
\newblock {\em ACM Computing Surveys}, 2022.

\bibitem{guo2022image}
Xiaojie Guo, Yang Yang, Chaoyue Wang, and Jiayi Ma.
\newblock Image dehazing via enhancement, restoration, and fusion: A survey.
\newblock {\em Information Fusion}, 86:146--170, 2022.

\bibitem{dcp}
Kaiming He, Jian Sun, and Xiaoou Tang.
\newblock Single image haze removal using dark channel prior.
\newblock {\em IEEE transactions on pattern analysis and machine intelligence},
  33(12):2341--2353, 2010.

\bibitem{he2016deep}
Kaiming He, Xiangyu Zhang, Shaoqing Ren, and Jian Sun.
\newblock Deep residual learning for image recognition.
\newblock In {\em Proceedings of the IEEE conference on computer vision and
  pattern recognition}, pages 770--778, 2016.

\bibitem{hdrnerf}
Xin Huang, Qi Zhang, Ying Feng, Hongdong Li, Xuan Wang, and Qing Wang.
\newblock Hdr-nerf: High dynamic range neural radiance fields.
\newblock In {\em Proceedings of the IEEE/CVF Conference on Computer Vision and
  Pattern Recognition}, pages 18398--18408, 2022.

\bibitem{hdrpleno}
Kim Jun-Seong, Kim Yu-Ji, Moon Ye-Bin, and Tae-Hyun Oh.
\newblock Hdr-plenoxels: Self-calibrating high dynamic range radiance fields.
\newblock In {\em Computer Vision--ECCV 2022: 17th European Conference, Tel
  Aviv, Israel, October 23--27, 2022, Proceedings, Part XXXII}, pages 384--401.
  Springer, 2022.

\bibitem{infonerf}
Mijeong Kim, Seonguk Seo, and Bohyung Han.
\newblock Infonerf: Ray entropy minimization for few-shot neural volume
  rendering.
\newblock In {\em CVPR}, 2022.

\bibitem{adam}
Diederik~P Kingma and Jimmy Ba.
\newblock Adam: A method for stochastic optimization.
\newblock {\em arXiv preprint arXiv:1412.6980}, 2014.

\bibitem{dpnerf}
Dogyoon Lee, Minhyeok Lee, Chajin Shin, and Sangyoun Lee.
\newblock Deblurred neural radiance field with physical scene priors.
\newblock {\em arXiv preprint arXiv:2211.12046}, 2022.

\bibitem{diffvolume}
Marc Levoy.
\newblock Efficient ray tracing of volume data.
\newblock {\em ACM Transactions on Graphics (TOG)}, 9(3):245--261, 1990.

\bibitem{li2020zero}
Boyun Li, Yuanbiao Gou, Jerry~Zitao Liu, Hongyuan Zhu, Joey~Tianyi Zhou, and Xi
  Peng.
\newblock Zero-shot image dehazing.
\newblock {\em IEEE Transactions on Image Processing}, 29:8457--8466, 2020.

\bibitem{aodnet}
Boyi Li, Xiulian Peng, Zhangyang Wang, Jizheng Xu, and Dan Feng.
\newblock Aod-net: All-in-one dehazing network.
\newblock In {\em Proceedings of the IEEE international conference on computer
  vision}, pages 4770--4778, 2017.

\bibitem{OTS}
Boyi Li, Wenqi Ren, Dengpan Fu, Dacheng Tao, Dan Feng, Wenjun Zeng, and
  Zhangyang Wang.
\newblock Benchmarking single-image dehazing and beyond.
\newblock {\em IEEE Transactions on Image Processing}, 28(1):492--505, 2019.

\bibitem{li2019semi}
Lerenhan Li, Yunlong Dong, Wenqi Ren, Jinshan Pan, Changxin Gao, Nong Sang, and
  Ming-Hsuan Yang.
\newblock Semi-supervised image dehazing.
\newblock {\em IEEE Transactions on Image Processing}, 29:2766--2779, 2019.

\bibitem{nerfacc}
Ruilong Li, Matthew Tancik, and Angjoo Kanazawa.
\newblock Nerfacc: A general nerf accleration toolbox.
\newblock {\em arXiv preprint arXiv:2210.04847}, 2022.

\bibitem{survey-dehaze}
Yu Li, Shaodi You, Michael~S Brown, and Robby~T Tan.
\newblock Haze visibility enhancement: A survey and quantitative benchmarking.
\newblock {\em Computer Vision and Image Understanding}, 165:1--16, 2017.

\bibitem{selective}
Xiao Liang, Runde Li, and Jinhui Tang.
\newblock Selective attention network for image dehazing and deraining.
\newblock In {\em Proceedings of the ACM Multimedia Asia}, pages 1--6. 2019.

\bibitem{griddehazenet}
Xiaohong Liu, Yongrui Ma, Zhihao Shi, and Jun Chen.
\newblock Griddehazenet: Attention-based multi-scale network for image
  dehazing.
\newblock In {\em Proceedings of the IEEE/CVF international conference on
  computer vision}, pages 7314--7323, 2019.

\bibitem{deblurnerf}
Li Ma, Xiaoyu Li, Jing Liao, Qi Zhang, Xuan Wang, Jue Wang, and Pedro~V.
  Sander.
\newblock Deblur-nerf: Neural radiance fields from blurry images.
\newblock {\em arXiv preprint arXiv:2111.14292}, 2021.

\bibitem{nerfw}
Ricardo Martin-Brualla, Noha Radwan, Mehdi S.~M. Sajjadi, Jonathan~T. Barron,
  Alexey Dosovitskiy, and Daniel Duckworth.
\newblock {NeRF in the Wild: Neural Radiance Fields for Unconstrained Photo
  Collections}.
\newblock In {\em CVPR}, 2021.

\bibitem{asm1976optics}
Earl~J McCartney.
\newblock Optics of the atmosphere: scattering by molecules and particles.
\newblock {\em New York}, 1976.

\bibitem{asm}
Earl~J McCartney.
\newblock Optics of the atmosphere: scattering by molecules and particles.
\newblock {\em New York}, 1976.

\bibitem{mildenhall2021nerf}
Ben Mildenhall, Pratul~P Srinivasan, Matthew Tancik, Jonathan~T Barron, Ravi
  Ramamoorthi, and Ren Ng.
\newblock Nerf: Representing scenes as neural radiance fields for view
  synthesis.
\newblock {\em Communications of the ACM}, 65(1):99--106, 2021.

\bibitem{asm2000chromatic}
Srinivasa~G Narasimhan and Shree~K Nayar.
\newblock Chromatic framework for vision in bad weather.
\newblock In {\em Proceedings IEEE Conference on Computer Vision and Pattern
  Recognition. CVPR 2000 (Cat. No. PR00662)}, volume~1, pages 598--605. IEEE,
  2000.

\bibitem{asm2002vision}
Srinivasa~G Narasimhan and Shree~K Nayar.
\newblock Vision and the atmosphere.
\newblock {\em International journal of computer vision}, 48(3):233, 2002.

\bibitem{regnerf}
Michael Niemeyer, Jonathan~T. Barron, Ben Mildenhall, Mehdi S.~M. Sajjadi,
  Andreas Geiger, and Noha Radwan.
\newblock Regnerf: Regularizing neural radiance fields for view synthesis from
  sparse inputs.
\newblock In {\em Proc. IEEE Conf. on Computer Vision and Pattern Recognition
  (CVPR)}, 2022.

\bibitem{volume}
Matthias Nie{\ss}ner, Michael Zollh{\"o}fer, Shahram Izadi, and Marc
  Stamminger.
\newblock Real-time 3d reconstruction at scale using voxel hashing.
\newblock {\em ACM Transactions on Graphics (ToG)}, 32(6):1--11, 2013.

\bibitem{quanti}
Tom Ouyang and Jack Tumblin.
\newblock Removing quantization artifacts in color images using bounded
  interval regularization.
\newblock {\em Northwestern University: Evanston, IL, USA}, 2006.

\bibitem{ffanet}
Xu Qin, Zhilin Wang, Yuanchao Bai, Xiaodong Xie, and Huizhu Jia.
\newblock Ffa-net: Feature fusion attention network for single image dehazing.
\newblock In {\em Proceedings of the AAAI conference on artificial
  intelligence}, volume~34, pages 11908--11915, 2020.

\bibitem{lolnerf}
Daniel Rebain, Mark Matthews, Kwang~Moo Yi, Dmitry Lagun, and Andrea
  Tagliasacchi.
\newblock Lolnerf: Learn from one look, 2022.

\bibitem{single}
Wenqi Ren, Si Liu, Hua Zhang, Jinshan Pan, Xiaochun Cao, and Ming-Hsuan Yang.
\newblock Single image dehazing via multi-scale convolutional neural networks.
\newblock In {\em Computer Vision--ECCV 2016: 14th European Conference,
  Amsterdam, The Netherlands, October 11-14, 2016, Proceedings, Part II 14},
  pages 154--169. Springer, 2016.

\bibitem{single2020}
Wenqi Ren, Jinshan Pan, Hua Zhang, Xiaochun Cao, and Ming-Hsuan Yang.
\newblock Single image dehazing via multi-scale convolutional neural networks
  with holistic edges.
\newblock {\em International Journal of Computer Vision}, 128:240--259, 2020.

\bibitem{dsnerf}
Barbara Roessle, Jonathan~T. Barron, Ben Mildenhall, Pratul~P. Srinivasan, and
  Matthias Nie{\ss}ner.
\newblock Dense depth priors for neural radiance fields from sparse input
  views.
\newblock In {\em Proceedings of the IEEE/CVF Conference on Computer Vision and
  Pattern Recognition (CVPR)}, June 2022.

\bibitem{tvloss}
Leonid~I Rudin, Stanley Osher, and Emad Fatemi.
\newblock Nonlinear total variation based noise removal algorithms.
\newblock {\em Physica D: nonlinear phenomena}, 60(1-4):259--268, 1992.

\bibitem{colmap}
Johannes~L Schonberger and Jan-Michael Frahm.
\newblock Structure-from-motion revisited.
\newblock In {\em Proceedings of the IEEE conference on computer vision and
  pattern recognition}, pages 4104--4113, 2016.

\bibitem{colmap2}
Johannes~L Sch{\"o}nberger, Enliang Zheng, Jan-Michael Frahm, and Marc
  Pollefeys.
\newblock Pixelwise view selection for unstructured multi-view stereo.
\newblock In {\em Computer Vision--ECCV 2016: 14th European Conference,
  Amsterdam, The Netherlands, October 11-14, 2016, Proceedings, Part III 14},
  pages 501--518. Springer, 2016.

\bibitem{dehazeformer}
Yuda Song, Zhuqing He, Hui Qian, and Xin Du.
\newblock Vision transformers for single image dehazing.
\newblock {\em arXiv preprint arXiv:2204.03883}, 2022.

\bibitem{nerfsr}
Chen Wang, Xian Wu, Yuan-Chen Guo, Song-Hai Zhang, Yu-Wing Tai, and Shi-Min Hu.
\newblock Nerf-sr: High quality neural radiance fields using supersampling.
\newblock In {\em Proceedings of the 30th ACM International Conference on
  Multimedia}, pages 6445--6454, 2022.

\bibitem{4KNeRF}
Zhongshu Wang, Lingzhi Li, Zhen Shen, Li Shen, and Liefeng Bo.
\newblock 4k-nerf: High fidelity neural radiance fields at ultra high
  resolutions.
\newblock {\em ArXiv}, abs/2212.04701, 2022.

\bibitem{zhang2021single}
Kuangshi Zhang and Yuenan Li.
\newblock Single image dehazing via semi-supervised domain translation and
  architecture search.
\newblock {\em IEEE Signal Processing Letters}, 28:2127--2131, 2021.

\bibitem{lpips}
Richard Zhang, Phillip Isola, Alexei~A Efros, Eli Shechtman, and Oliver Wang.
\newblock The unreasonable effectiveness of deep features as a perceptual
  metric.
\newblock In {\em Proceedings of the IEEE conference on computer vision and
  pattern recognition}, pages 586--595, 2018.

\bibitem{hierarchical}
Xiaoqin Zhang, Jinxin Wang, Tao Wang, and Runhua Jiang.
\newblock Hierarchical feature fusion with mixed convolution attention for
  single image dehazing.
\newblock {\em IEEE Transactions on Circuits and Systems for Video Technology},
  32(2):510--522, 2021.

\bibitem{cap}
Qingsong Zhu, Jiaming Mai, and Ling Shao.
\newblock A fast single image haze removal algorithm using color attenuation
  prior.
\newblock {\em IEEE transactions on image processing}, 24(11):3522--3533, 2015.

\end{thebibliography}
}

\end{document}